%% file: bmvc_final.tex
\documentclass{bmvc2k}

\title{Cross-Modal Generative Augmentation for Visual Question Answering}
\addauthor{Zixu Wang}{zixu.wang@imperial.ac.uk}{1}
\addauthor{Yishu Miao}{y.miao20@imperial.ac.uk}{1}
\addauthor{Lucia Specia}{l.specia@imperial.ac.uk}{12}

\addinstitution{
 Department of Computing\\
 Imperial College London\\
 London, UK
}
\addinstitution{
 Department of Computer Science
 University of Sheffield\\
 Sheffield, UK
}

\runninghead{Wang, Miao, Specia}{Cross-Modal Generative Augmentation for VQA}
        
\def\eg{\emph{e.g}\bmvaOneDot}

\usepackage{tabularx,ragged2e}
\usepackage{multicol}
\usepackage{multirow}
\usepackage{booktabs}
\newcommand*{\ie}{\textit{i.e.}}

\begin{document}

\maketitle

\begin{abstract}
Data augmentation has been shown to effectively improve the performance of multimodal machine learning models.
This paper introduces a generative model for data augmentation by leveraging the correlations among multiple modalities. 
Different from conventional data augmentation approaches that apply low-level operations with deterministic heuristics,
our method learns a generator that generates samples of the target modality conditioned on observed modalities in the variational auto-encoder framework. 
Additionally, the proposed model is able to quantify the confidence of augmented data by its generative probability, and can be jointly optimised with a downstream task. 
Experiments on Visual Question Answering as downstream task demonstrate the effectiveness of the proposed generative model, which is able to improve strong UpDn-based models to achieve state-of-the-art performance.
\end{abstract}

%-------------------------------------------------------------------------
\section{Introduction}
Multimodal machine learning is a multidisciplinary field combining language, vision, and speech processing to address a multitude of tasks \cite{Ngiam2011MultimodalDL, 8269806, Mogadala2019TrendsII, Uppal2020EmergingTO, bisk-etal-2020-experience}.
However, a major bottleneck in multimodal learning is the need for multi-way parallel data, i.e. data with all modalities for all samples.
For example, Visual Question Answering (VQA) \cite{NIPS2016_9dcb88e0, YangHGDS16, Anderson_2018_CVPR, pythia18arxiv, tan-bansal-2019-lxmert, Su2020VL-BERT:, chen2020uniter, gan2020large} requires learning from parallel data across three different sources -- image, question and answer, a costly resource if created  
%it is non-trivial to annotate images with multiple reasonable question answer pairs
at large scale.
In this paper we propose a generative model that leverages the joint distribution of multiple modalities from existing datasets to carry out data augmentation for VQA.
% More specifically, we construct an generator to generate meaningful question-answer (QA) pairs for images to improve VQA models. 

%\paragraph{VQA, VQA requires data augmentation} Visual Question Answering (VQA) \citep{VQA2015} has become one of the main challenges in multimodal machine learning which requires effective learning from data across multiple modalities (\ie image and textual question) and is essential of machine comprehension and reasoning \citep{NIPS2016_9dcb88e0, conf/cvpr/YangHGDS16, Anderson_2018_CVPR, pythia18arxiv, tan-bansal-2019-lxmert, Su2020VL-BERT:, chen2020uniter, gan2020large}. 

% The joint distribution of the three modalities is able to provide rich cross-modal information to help us recover one modality by conditioning on other ones.
% Without massive human labor for annotation, our proposed approach can effectively make use of the existing data distribution to boost the baseline models with augmented data from labelled or unlabelled resources.
% \begin{figure}[t]
%   \centering
%   \includegraphics[width=0.5\textwidth]{figures/illustration}
%   \caption{Illustration of the proposed generative process for data augmentation. The top part describes the conventional VQA pipeline consisting of an image and annotated QA pairs. The bottom part is our approach to generating QA pairs for data augmentation.}
%   \label{fig:illustration}
% \end{figure}

\begin{figure}[t]
\centering
\begin{tabular}{cc}

\includegraphics[width=6.15cm, height=2.2cm]{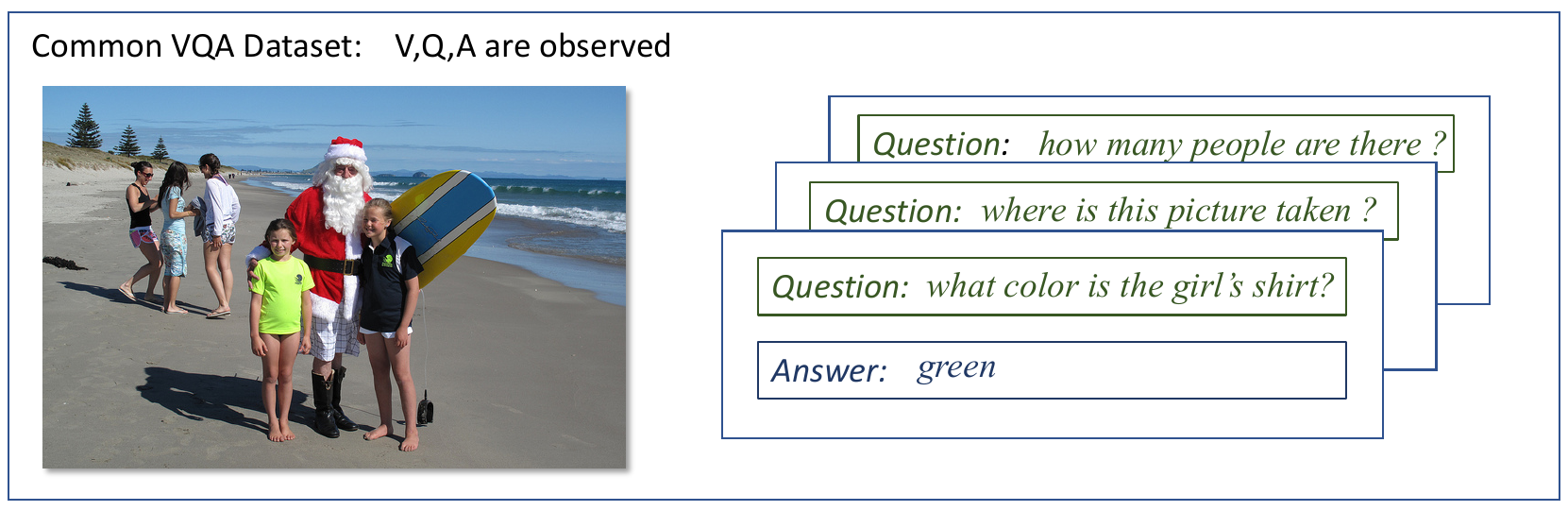}
\includegraphics[width=6.15cm, height=2.2cm]{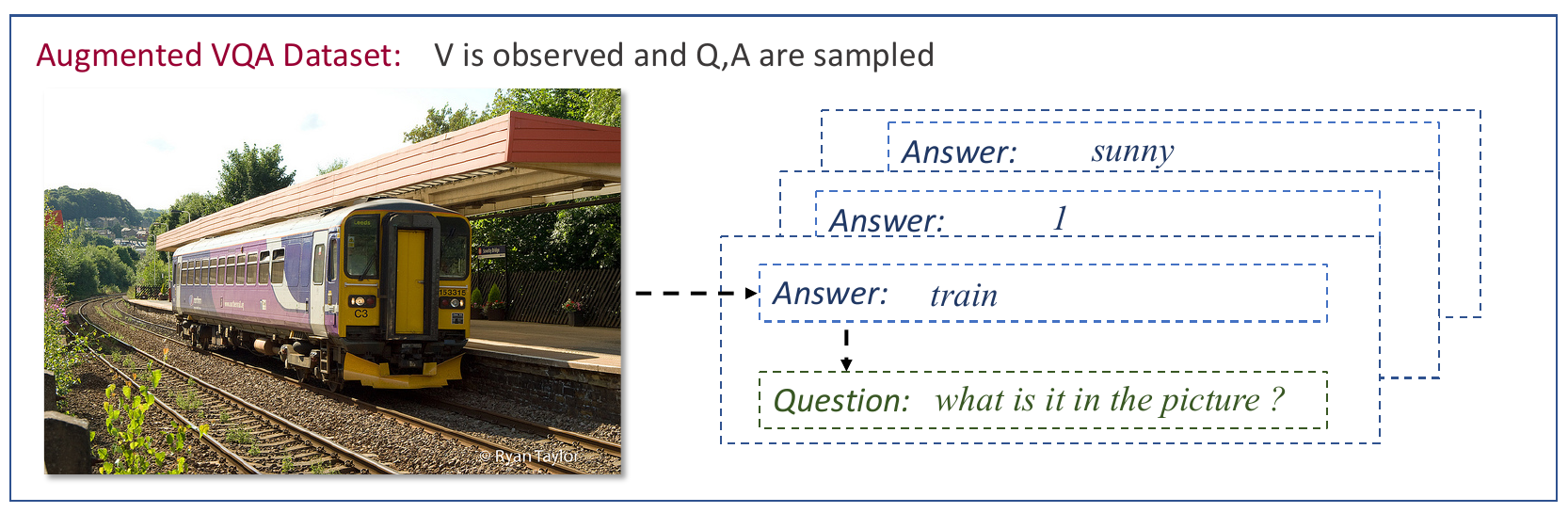}

\end{tabular}
\caption{Illustration of the proposed generative process for data augmentation. The left one describes the conventional VQA pipeline consisting of an image and annotated QA pairs. 
The right one is our approach to generating QA pairs for data augmentation. 
% where possible answers are sampled from a given image followed by question generation conditioned on the image and a sampled answer.
}
\label{fig:illustration}
\end{figure}

% Previous research on data augmentation for VQA suffers from two major challenges.
There are two major challenges for data augmentation.
First, the augmented data should contain meaningful \textbf{variations} and less repetition, and second, the \textbf{reliability} of the augmented data should be guaranteed and effectively evaluated.
The repetitive, insignificant variants or unreliable augmented data could have a negative effect on  downstream tasks. 
Conventional data augmentation approaches apply low-level operations, e.g. adding noise or using replacements via heuristics.
For example, \citet{chen2020counterfactual} and \citet{gokhale-etal-2020-mutant} apply visual and semantic transformations which create additional counterfactual samples to improve the model's sensitivity to trivial noise in VQA. 
This provides enough variations but requires specific design and extensive preprocessing to fit the task domain.
% Besides, it lacks valid approaches to quantify the reliability of augmented data samples.

Our proposed model explores the generative abilities of conditional distributions for augmenting question-answer (QA) pairs given only images.
Figure \ref{fig:illustration} shows the difference between a conventional VQA model and our proposed generative framework. The annotated QA pairs are directly used in supervised learning to predict answers in a VQA model, while our model aims at generating QA pairs from unlabelled images.
We first learn the conditional distributions of different modalities using annotated images and QA pairs. 
Then, we introduce Q and A as two discrete latent variables, and construct a generative model for unlabelled images to generate QA pairs as augmented data.
% Finally, the augmented data is used in the downstream VQA task to boost the performance of base models.  
Finally, the augmented data with reliability scores are used for training to improve on strong VQA base models.  
Without any specific low-level operations on the data, we utilise the dynamics of the generative distributions and the compositionality of multiple modalities to explore the novel QA pairs given an image.
Additionally, the generators are optimised by the REINFORCE algorithm \citep{10.1007/BF00992696} while minimising the variational lower bound.
The generative approach has the following advantages.
First, the model has strong generalisation ability to incorporate additional unseen and unlabelled images.  
Diverse QA pairs can be generated by exploring the dynamics of the generator and the compositionality of multiple modalities.
%-- i.e. without ground truth questions and answers. 
Second, the answers are sampled from images before generating the questions, hence it is less prone to exploit linguistic priors in questions and to generate trivial QA pairs that are irrelevant to the given images.
Third, the augmented data can be quantified by the generative distribution, which acts as reliability scores of QA pairs for downstream VQA training.
Our approach opens a promising new direction for multimodal learning, with strong potential for cross-modal understanding and generalisation.

Using VQA as downstream  task, our approach outperforms base models (UpDn and LXMERT),   and substantially improves strong UpDn models, leading to the state-of-the-art performance on the task.

%Our {\bf main contributions} are as follows:
%\begin{itemize}
%    \item We propose a generative model to learn a generator for producing QA pairs given images (Section \ref{sec:generative}).
    % Under the general variational auto-encoder framework, the model can effectively learn the cross-modal distributions.
%    \item We validate our model on the downstream VQA task. The generator is used to infer missing QA pairs with high correlation with the given image to improve answering performance. This is especially beneficial to generate QA pairs for unlabelled images for large-scale training (Section \ref{sec:vqa}). 
    % \item We introduce a generative framework by first predicting possible answers from an image directly, and propose a conditional language model to generate questions corresponding to the answers and the image. To the best of our knowledge, this is the first attempt to use generative models in the VQA task.
%    \item The results of our experiments (Sections \ref{sec:results})
    % \todo{add references - both here?} 
%    on the benchmark dataset VQA-CP v2 indicate that our approach outperforms prior work for VQA and boosts the strong UpDn models to the state-of-the-art performance. 
%\end{itemize}

\begin{figure*}[t]
  \centering
  \includegraphics[width=0.90\textwidth]{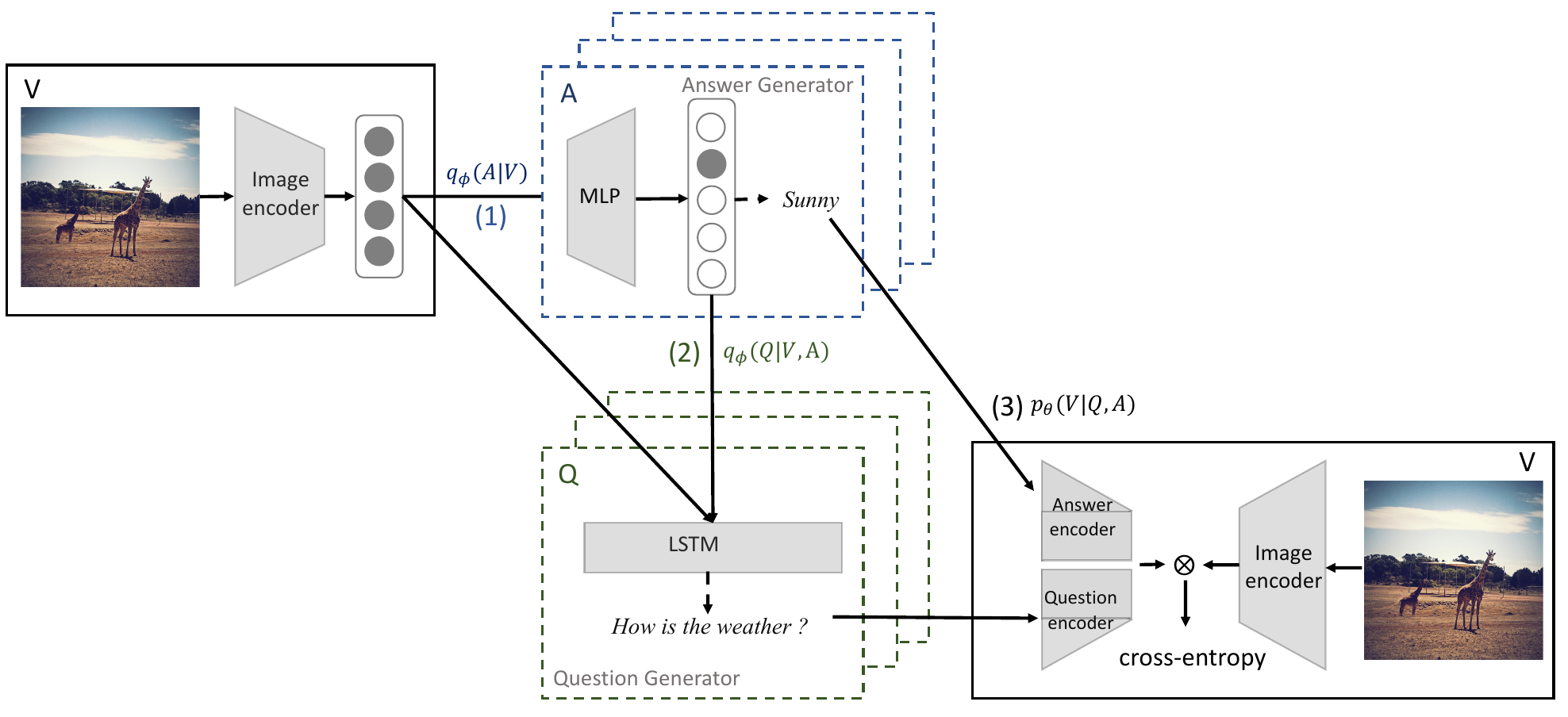}
  \caption{
  Architecture of the proposed generative model for data augmentation in the VQA classification task.
%   We combine the \textbf{generator} and \textbf{joint cross-modal distribution} for simplicity.
  For the training of the generator $q_{\phi}(Q, A|V)$, 
  (1) given an image, we first build an image classification module $q_{\phi}(A|V)$ to predict possible answers; 
  (2) with a sampled answer (\eg ``sunny'') from answer distribution, we build a conditional question generator $q_{\phi}(Q|V,A)$ using an LSTM to generate questions (\eg ``how is the weather ?'') corresponding to the image and the sampled answer;
%   For the cross-modal pretraining on the joint distribution of (image, question, answer) $p_{\theta}(V, Q, A)$,
%   we first sample an answer according to $p_{\theta}(A)$; then use $p_{\theta}(Q|A)$ generating a question given the answer, 
then (3) we apply a neural network $p_{\theta}(V|Q, A)$ to measure the score of relevance between generated QA pairs and the image.
%   On the upper right, we use the generated QA paired with the image for VQA training.
  }
%   \todo{are the letters (a-c) in the caption of figure 2 supposed to be mapped to parts of the image? if not maybe remove them (just the letters themselves)}
  \label{fig:model}
\end{figure*}

\section{Related Work}
\paragraph{Data Augmentation in VQA}
There has been extensive work on improving VQA robustness with data augmentation \citep{ tang2020semantic, ijcai2020-151, kafle-etal-2017-data,  gokhale-etal-2020-mutant, Tang2020SemanticEA, ray-etal-2019-sunny}. 
The pioneer work is introduced by \citet{kafle-etal-2017-data} where they generate new questions by using semantic annotations
on images. 
\citet{meet2019cycle} proposes a cycle consistency training scheme where it generates questions and trains the model with question and answer consistency.
Augmenting VQA training data with counterfactual samples has also been proven effective \citep{Agarwal2020TowardsCV, Kant2020ContrastAC} with complementary samples by masking critical objects in images or words in questions, and assigning different ground-truth answers (\eg, CSS \citep{chen2020counterfactual} and MUTANT \citep{gokhale-etal-2020-mutant}).
However, most data augmentation work focuses on explicitly injecting semantic and visual perturbations on raw inputs or using adversarial samples to provide contrastive learning signals.
Our work augments the original dataset with implicitly generated QA pairs especially for unlabelled images. This helps the model learn correlations between modalities.
% \todo{is it worth adding the method names here if we have them in our Tables of results? otherwise it will look strange to think we don't compare against them - e.g. MUTANT}

\paragraph{Visual Question Generation}
%In contrast to answering questions, 
Generating questions from images has received equal interests as VQA \citep{mostafazadeh-etal-2016-generating, Jain_2017_CVPR, Li_2018_CVPR, lee-etal-2020-generating, Xiong_2020_CVPR, vedd2021guiding}. 
Several recent works have explored the task of visual question generation with variational auto-encoders and maximising mutual information with answer categories \citep{Jain_2017_CVPR, krishna2019information}. 
Our work draws inspiration from VQG by incorporating a generation module in the generative framework.
% We believe that with question generator our model can learn better interaction and correlations between images and questions.

% In order to measure the correlations between generated QA pairs and the image, 
Our generative model incorporates a cross-modal retrieval module  \citep{Gu_2018_CVPR, Wang_2019_ICCV} to measure the correlations between generated QA pairs and the image.
% which has been a hot research topic in both computer vision and natural language processing communities \citep{Gu_2018_CVPR, Wang_2019_ICCV}.
Cross-modal retrieval performance relies on appropriate representations for multi-modal data.
Most of the existing studies on cross-modal retrieval mainly focus on learning a high-level common space and exploiting visual-semantic embedding to calculate the similarities between image and sentence features with ranking loss \citep{socher-etal-2014-grounded, kiros2014unifying, Vendrov2016OrderEmbeddingsOI, Wang_2016_CVPR, Fang_2015_CVPR}. 
We build a classifier to output the reliability scores of generated QA pairs and images.

\section{Model}
\label{sec:generative}
Variational auto-encoders (VAE) are generally applied for unsupervised representation learning of language and images \citep{kingma2013auto, pmlr-v32-rezende14, pmlr-v32-mnih14, NIPS2014_d523773c, pmlr-v48-miao16, miao-blunsom-2016-language}.
Inspired by the idea of VAE for data augmentation \citep{pmlr-v70-hu17e, Hou2018SequencetoSequenceDA, yoo2019data}, we propose a generative framework in order to explore cross-modal interactions in the multimodal data for the task of VQA.

% \textcolor{red}{VQA loss, Generative augmentation VQA}

Here, $V$, $Q$, $A$\ are used to denote the input image, question, and the answer respectively.
Figure \ref{fig:model} shows the structure of variational auto-encoder for VQA. $Q$, $A$ are the two latent variables introduced to represent questions and answers, respectively.
The training objective consists of the variable lower bound and the cross-entropy (CE) loss for VQA classification:
\begin{equation}
\label{eq:objective}
    E_{q_{\phi}(Q,A|V)} [\log p_{\theta}(V, Q, A) - \log q_{\phi}(Q, A|V)] + p_{\psi}(A|V,Q)
\end{equation}
where $q_{\phi}(Q,A|V)$ is the generator that generates corresponding QA pairs for given images and can be factorised into answer generator $q_{\phi}(A|V)$ and question generator $q_{\phi}(Q|V,A)$;
$p_{\theta}(V,Q,A)$ is the cross-modal distribution that regularises the QA samples, and $p_{\psi}(A|V,Q)$ is the multi-label classification loss for VQA.
% Generally, we can further decompose the $p_{\theta}(V,Q,A)$ into \textbf{prior} $p_{\theta}(Q, A)$ and \textbf{decoder} $p_{\theta}(V|Q,A)$. 
Note that the lower bound will act as a confidence measure of generate QA pairs and will be used to reweigh the cross-entropy loss while training the downstream VQA model.

The overall architecture of our generative model is illustrated in Figure \ref{fig:model}.
The training steps are summarised as follows: 
the joint distribution of three modalities (V, Q, and A) is modelled to learn prior knowledge on cross-modal interactions; 
optimise the generator to generate QA pairs by first predicting answers from the image and then generating question;
sample QA pairs for additional unlabelled images to assist large-scale training for VQA.
% \begin{enumerate}
% \vspace{-0.1cm}
% \item The joint distribution of three modalities (V, Q, and A) is modelled to learn prior knowledge on cross-modal interactions;
% \vspace{-0.1cm}
% \item Optimise the generator to generate QA pairs by first predicting answers from the image and then generating question;
% \vspace{-0.1cm}
% \item Sample QA pairs for additional unlabelled images to assist large-scale training for VQA.
% \end{enumerate}

% \todo{from here on, in each subsection, should we refer to the numbers (1-3) in figure 2 for clarity}
\subsection{Generative Learning}
The first term in the training objective is the variational lower bound that is optimised for learning cross-modal distributions.
In order to construct this lower bound, we need to model both the generative distribution $q_{\phi}(Q,A|V)$ and the cross-modal distribution $p_{\theta}(V, Q, A)$.

% \zw{I have checked the content in 2.2. it is good so far i think. can polish later.}
\subsubsection{Generative Distribution}
The training of the generator $q_{\phi}(Q,A|V)$ can be decomposed into:
\begin{equation}
    q_{\phi}(Q,A|V) = q_{\phi}(Q|V, A) q_{\phi}(A|V)
\end{equation}
Specifically, given a triplet ($V, Q, A$), we build an answer generator $q_{\phi}(A|V)$ and a conditional question generator $q_{\phi}(Q|V, A)$.
The original image $V$ and the generated $Q$ are then combined to obtain an answer prediction $A$ using the conventional VQA model $p_{\psi}(A|V, Q)$.  

The design of our generative components is based on two hypotheses. 
First, a model that can predict possible answer candidates directly from a given image has better understanding of the visual content and the dependency between the image and the answers. 
Second, assuming the predicted answers have high correlation with the image, the conditional language model can generate valid questions considering both the image and the predicted answers.

\paragraph{Answer Generator - $q_{\phi}(A|V)$}
% \paragraph{Image Processor} 
As illustrated in Figure \ref{fig:model} step (1), first we need to build an image classification model to predict possible answers to a given image.
Following typical setups in VQA models, we encode the input images as regional visual feature representations. 
Specifically, these features are extracted from a bottom-up approach \citep{Anderson_2018_CVPR}, where the input image is passed through a ResNet CNN within the Faster R-CNN framework to obtain a vector representation.
We take pretrained visual features as a preprocessing step for efficiency, and keep them fixed during the training of the VQA model. 
We follow the same setup for both labelled and unlabelled images in our experiments.
% and pre-trained on the MSCOCO 2015 dataset~\citep{lin2014microsoft}. 
% We represent each input image as extracted from the Faster R-CNN~\citep{NIPS2015_5638} pre-trained on Visual Genome~\citep{krishna2017visual}. 
% As in the standard VQA model using an attention mechanism, attention from the question features to the image features is usedto output a weight for the feature vector at each spatial position in the feature map; which is first normalised and then used for performing a weighted sum over the spatial positions to produce a single feature vector to represent the image. 
% \paragraph{Generator}

In our generative framework, the predicted answer distributions given images can be modelled as $q_{\phi}(A|V)$. 
We build the classifier by using a non-linear fully connected layer and batch normalisation on the pretrained image features, followed by a projection to the space of all possible answers and a softmax layer.
This can be seen as a multi-label image classification task where the labels are possible VQA answers corresponding to the given image instead of  image classification categories. 
Without seeing the questions with strong language priors, the model can directly learn the mapping from the image to answer candidates.

% Note that the answer classifier works differently on labelled and unlabelled images. 
% The labelled images refer to the data from original VQA dataset with annotated QA pairs.
% For a labelled image, we take the answers in multiple QA pairs corresponding to the image as ground-truth labels and pretrain the classifier. 
% For an unlabelled new image which we use for training downstream VQA model, we use the prediction from the pretrained classifier to obtain an answer distribution for the image.
% We then sample possible answers from this distribution for guiding the question generation.

\paragraph{Question Generator - $q_{\phi}(Q|V, A)$}
The second stage of our generative process is to generate questions conditioned on the given image and sampled answers, as shown in Figure \ref{fig:model} step (2).
The goal of our question generator is to define a conditional language model $q_{\phi}(Q|V, A)$ to learn the transformations from one modality (\ie image) to another (\ie question) with the help from possible answers.
This is similar to a visual question generation (VQG) task that is aimed at generating not only relevant but diverse questions to each image. 
However, typical VQG models only take images as input, and thus are not goal-driven and do not guarantee that the generated question corresponds to a specific type of answer. 
We follow the common setup in \citep{meet2019cycle, krishna2019information} to encode the answer along with the image before generating the question. Such an approach allows the model to condition its question on the answer.

We model the question generator by an image encoder, answer encoder, and an LSTM decoder.
With predicted answer distributions, the model can generate various plausible questions for the sampled answers.
The image encoder transforms the attended image features to lower dimensional feature vectors, and the answer encoder takes the distribution over the answer space as input and outputs the vector representation of the answer.
The image and answer representation are then fused together and passed through the LSTM decoder to generate a question, a process that is optimised by minimising the negative log likelihood with teacher-forcing. 

% Now we have obtained the generative distribution $q_{\phi}(Q,A|V) = q_{\phi}(Q|V, A)q_{\phi}(A|V)$ for both labelled and unlabelled images.
% For unlabelled images, the advantage is that we can generate potentially valid questions without additional supervision but only the images and the answer distribution predicted from the images. 

\subsubsection{Cross-Modal Distribution}
We need to model the cross-modal distributions on the existing VQA dataset, where all the modalities are observed.
The cross-modal distribution can be decomposed into:
\begin{align}
    p_{\theta}(V, Q, A) =& p_{\theta}(V|Q, A)p_{\theta}(Q,A) \nonumber \\ 
                        =& p_{\theta}(V|Q, A)p_{\theta}(Q|A)p_{\theta}(A)
\end{align}

We first learn to model the joint distribution of QA pairs.
We build the model to learn the prior knowledge of QA pairs $p_{\theta}(Q, A)$ through $p_{\theta}(Q|A)p_{\theta}(A)$. 
Firstly, the prior distribution of answer candidates $p_{\theta}(A)$ is obtained directly from the dataset.
With a sampled answer from the prior distribution, we build a question generator conditioned on the answer.
Note that instead of obtaining a high quality generative distribution for QA pairs, we learn a joint distribution of the two modalities as the prior knowledge of QA pairs.

The QA pairs are expected to have high dependency and correlation with the given image. 
However, not all the QA pairs are coherent and consistent with the image, which leads to negative noises during training. 
Inspired by \citet{meet2019cycle}, we overcome this issue by modelling the conditional distribution of images given QA pairs.
As shown in Figure \ref{fig:model} part (3), we build a neural network to model $p_{\theta}(V|Q, A)$ and score the correlation/relevance between each QA pair and the image.
We add a layer to fuse question and answer representations together, then multiply the fused QA vector with the image representation through another fusion layer.
Having the representations of both QA pairs and images, we feed them into a binary classifier to output a relevance score of the image and QA pairs after a sigmoid layer.

\subsection{VQA Classification}
\label{sec:vqa}
Following the conventional setup in VQA, the image feature $V$ and question representation $Q$ are extracted from the image encoder and the question encoder.
The VQA task is constructed as a classification problem to output the most likely answer $A$ from a fixed set of answers based on the content of the image $V$ and question $Q$.
Following \citet{teney2018tips} and \citet{ijcai2020-151}, instead of softmax we use sigmoid outputs in our VQA training to cast it as a multi-label training objective:
\begin{equation}
     p_{\psi}(A|V, Q) = \rm{CE}\ (f_{\rm{VQA}}(V,Q), A)
\end{equation}
where \rm{CE}\ represents the cross-entropy loss in VQA models.

% VQA models tend to capture language biases and learn spurious correlations between question and answers.
% For example, to answer the questions \textit{``How many XXX?"}, the model can predict \textit{``2"} based on simple correlations in the dataset without considering the image. 
% Inspired from \citep{ijcai2020-151}, we replace the original images in the triplets of (image, question, answer) by a randomly selected image from the dataset to create negative samples. 
% These new samples are labelled as irrelevant and trained by minimising the confidence of
% irrelevant pairs being relevant, that is minimising $p_{\psi}(A|V, Q)$ directly for these irrelevant pairs.
% This mechanism realises data augmentation by adding irrelevant samples such that each question is paired with equal amounts of relevant and irrelevant images. 
% This can teach the model to learn the correlation between true tuples of image and the QA pairs, which can improve the model sensitivity to the visual content instead of superficial linguistic information.

% \paragraph{Reweighted Augmented Data Loss}
\begin{equation}
\label{eq:reweighing}
    p_{\psi}(A|V, Q) = \rm{CE}\ (f_{\rm{VQA}}(V,Q), A) * \mathcal{R}
\end{equation}

One of the assumptions of our proposed generative training scheme is that the generated QA pairs are always semantically and syntactically correct and have high correlation with the image. 
However, this is not always the case. 
In order to overcome this issue, we propose a reweighing mechanism based on a reliability score obtained from our generative objective, as in Equation \ref{eq:reweighing}.
We define the reliability and confidence of each generated QA pair as $\mathcal{R}$ using the generative objective in Equation \ref{eq:objective} and use it to reweigh the cross-entropy loss during VQA training. 
We add this reliability score to make the model aware of the confidence and quality of the augmented samples.

\section{Experimental Setup}
\label{sec:experiments}
%\subsection{Experimental setup}
\subsection{Datasets}
We train and evaluate our models on the VQA-CP-v2 \citep{Agrawal_2018_CVPR} and VQA-v2 \citep{VQA2015} datasets. 
These are the most commonly used VQA benchmarks, where VQA-CP-v2 is created from VQA-v2 to evaluate robustness and generalizability of VQA models, by reorganising the train and validation splits.
% In VQA-CP-v2, language priors cannot be relied upon for a correct prediction because the QA pairs in the training set and test set have different distributions. 
We also report the results of our model on the validation split of VQA-v2, which contains a strong language prior. VQA-CP-v2 contains 121K images, 438K questions, and 4.4M answers for training and 98K images, 220K questions and 2.2M answers for testing. 
We evaluate our model on the validation split of VQA-v2, which contains 83K images, 444K questions, and 4.4M answers for training and 41K images, 210K questions and 2.1M answers for validation.

The unlabelled images come from the Visual Genome (VG) \citep{DBLP:journals/ijcv/KrishnaZGJHKCKL17} and MS COCO 2017-unlabelled \citep{lin2014microsoft} datasets. 
We directly use the images from these datasets and generate QA pairs relating to the images using proposed generator. 
The number of additional images is 101K from VG and 120K from MS COCO 2017-unlabelled data, which are combined to provide 221K unlabelled images in total. 
Note that we do not filter out VG images that overlap with original VQA dataset as our generator can generate new QA pairs relating to these images.

\subsection{Benchmark Models}
% \todo{I suggest changing the name of this subsection to "Base models" and using that term in the text too - baseline gives the impression of weak models}
Most prior works on VQA, such as the ones we compared our model against -- GVQA \citep{Agrawal_2018_CVPR}, RUBI \citep{NEURIPS2019_51d92be1}, SCR \citep{NEURIPS2019_33b879e7}, LMH \citep{clark-etal-2019-dont}, CSS \citep{chen2020counterfactual} -- are all built based on UpDn \citep{Anderson_2018_CVPR}. 
Therefore, we also build our generative model using UpDn \citep{Anderson_2018_CVPR} as its backbone and investigate the efficacy of the extension under the generative paradigm. 
UpDn incorporates bottom-up attention in VQA by extracting features associated with image regions proposed by Faster RCNN \citep{NIPS2015_14bfa6bb} trained on Visual Genome \citep{DBLP:journals/ijcv/KrishnaZGJHKCKL17}.
% UpDn model won the VQA Challenge in 2017 \citep{teney2018tips} and achieves 66.25\% accuracy on VQA-v2 test-dev. 
We use the evaluation code from official VQA challenge \citep{VQA2015}.

% \paragraph{Self-Supervised / negative} 
% \zw{to be rephrased}
% As a question can only be answered when the given image containing the key information for answering the question, it is necessary to estimate whether the given question and image are relevant or not before answering the question. 
% \citep{ijcai2020-151} introduce an auxiliary task named question-image correlation estimation to estimate the relevance between questions and images. 
% Specifically, the model first automatically generates a set of balanced question-image pairs with binary labels (relevant and irrelevant), which are then consumed by the self-supervised auxiliary task to assist the VQA model to overcome language priors. They incorporate the auxiliary task into the VQA model by feeding the relevant and irrelevant pairs. 
% When fed a relevant question-image pair, the model is encouraged to predict the correct answer with a high confidence score, where the confidence score is the probability of the question-image pair being relevant. 
% On the contrary, the model is pushed to predict the correct answer with a low confidence score when the input pair is irrelevant.

\subsection{Implementation Details}
% The proposed generative VQA model is a model-agnostic training scheme that can be incorporated into different VQA architectures. 
% In this paper, we mainly evaluate our method based on UpDn \citep{Anderson_2018_CVPR}.
% and also report the results using LXMERT \citep{tan-bansal-2019-lxmert} (a unified vision-language pretraining framework based on BERT \citep{devlin-etal-2019-bert}) for the generalisation ability. 
Following previous work, we take the pretrained image features extracted from a ResNet CNN within a Faster R-CNN framework. 
Each image is transformed into a $K x 2048$ dimensional vector representation where $K=36$ in our setup represents a set of objects in the image.
Each question is trimmed into a sequence with a maximum length of 14 words and initialised using the 300-dimensional word embeddings from GloVe \citep{pennington-etal-2014-glove}. 
A sentence-level representation is obtained by feeding the word embeddings into a GRU \citep{chung2014empirical} with 1280 dimensions. 
While using sampled answers to guide question generation, we add an answer encoder to transform the samples answer to a representation of dimension 512.

We pre-train the model with the proposed generative objective for 10 epochs ($\sim$4 hours) and fix it for generating QA pairs to assist downstream VQA training. 
This number of epochs is chosen because our generative model converges and the quality of generated questions does not improve after these epochs of training.
The number of parameters is 90M and the model is trained on 2 RTX 2080Ti.
We use batch size of 256 and adapt the Adam optimizer \citep{kingma2014adam} with the initial learning rate of 0.001.

\section{Results}
\label{sec:results}
In this section we present our experimental results including quantitative and qualitative analysis.
We first compare the results with previous approaches including strong baseline models and SOTA benchmarks;
then we validate the efficacy of each component proposed in our framework in ablation studies;
finally we provide a qualitative analysis of the generated QA pairs. 

\input{tables/tab_results}

%\subsection{Ablative Studies}
\subsection{Comparison with Alternative Approaches}
In Table \ref{tab:results}, we compare our proposed framework with previous SOTA approaches on two benchmarks: VQA-CP-v2 and VQA-v2. 
We compare our generative VQA model against existing models. 
RUBi\citep{NEURIPS2019_51d92be1}, SCR\citep{NEURIPS2019_33b879e7}, LMH\citep{clark-etal-2019-dont}, and CSS\citep{chen2020counterfactual} are built on UpDn\cite{Anderson_2018_CVPR} by adding different de-biasing components to mitigate superficial language biases \cite{niu2020counterfactual} and improve robustness of VQA models.
Besides, CSS\citep{chen2020counterfactual} and MUTANT\citep{gokhale-etal-2020-mutant} make use of data augmentation to provide large-scale data sizes for VQA training.
% \todo{can you say a little more about this? what is the expected effect of debiasing components?}

We show that our generative VQA model outperforms most of the alternative approaches above.
For VQA-CP-v2, our method achieves 60.70 accuracy on all question types, competitive with the top results from MUTANT.
% Among previous approaches, CSS and MUTANT make use of data augmentation to provide large-scale data sizes.
% However, different from our generative framework, they rely on costly visual and semantic transformation for augmenting the data,
% \todo{worth saying something more negative about them, or else why do we want to get rid of it? costly? also, should we say their approach is a pipeline where data augmentation is done independently of the downstream task?} 
% which has lower generalisation ability. 
% \textcolor{red}{CSS and MUTANT - introduce first - and midrow in table: also use data augmentation}
Our method shows improvements with 3.20 on the Yes-No category, 15.52 on Number-based questions; also 0.83 compared to MUTANT on the Yes-No category.
Our model improves substantially on ``Num'' questions.
When built on UpDn and evaluated on the VQA-v2 val dataset, our framework achieves better performance (\%64.09) than UpDn and outperforms MUTANT (\%62.56).
This indicates that our generative augmentation model has higher generalisation ability to different VQA datasets, compared to MUTANT.

% To compare with the state-of-the-art model MUTANT (which is built using LXMERT as the backbone) evaluated on VQA-CP-v2, we further train our model based on LXMERT, where our generated QA pairs are used to fine-tune the pre-trained multimodal model.
% The results show that our model reaches some improvement (46.93\%) over LXMERT (46.23\%), which further shows the efficacy of generated QA pairs. 
% This marginal improvement might be due to less contrastive samples that our model generates, compared to MUTANT.
% The augmented data in MUTANT is from manually introduced mutations such as removing object instances and colour inversion for a strong contrastive learning signal.
% Although MUTANT is also focused on data augmentation and achieves significant improvements on the pre-trained LXMERT ($46.23\% \rightarrow 59.69\%$), they rely on low-level operations with deterministic heuristics and apply costly extensive transformations on original samples independently of VQA training. We believe this is intentionally designed for this dataset and limits the generalisation ability.

% my other take
To further prove the effectiveness of our generative model, we conduct experiments based on pre-trained visual-language framework - LXMERT \citep{tan-bansal-2019-lxmert}, which is also the backbone of the state-of-the-art model on VQA-CP-v2 - MUTANT \citep{gokhale-etal-2020-mutant} using data augmentation.
Specifically, our generated QA pairs are used to fine-tune the pre-trained LXMERT model.
Our model reaches some improvement over LXMERT on VQA-CP-v2, which further shows the efficacy of our generated QA pairs.
Although MUTANT is also focused on data augmentation and achieves significant improvements over LXMERT, their augmented data is from manually introduced mutations such as removing object instances and colour inversion for a strong contrastive learning signal.
The augmentation relies on low-level operations with deterministic heuristics and applies costly extensive transformations on original samples independently of VQA training.
We believe this is intentionally designed for VQA-CP-v2 dataset and limits the generalisation ability, while our goal is to introduce a generative approach that can be generalised to different multimodal tasks for cross-modal data augmentation.

% It is worth noting that we use the generative model to incorporate unlabelled images and generate QA pairs to be added to the training set, but this operation is not present in the test set.

% Particularly, it improves the performance of UpDn with a 0.61\% performance gains (64.09\% vs. 63.48\%). 
% For completeness, we further compared the performance drop between the two benchmarks.
% Different from previous models that suffer severe performance drops (e.g., 23.74\% in UpDn, and 9.19\% in LMH), our model can significantly decrease the performance drop into 4.15\%, which demonstrate that the effectiveness to further reduce the language biases in VQA.

\subsection{Evaluation of Model Components}
In order to evaluate the efficacy of each component of our generative model, we conduct an ablation study. Table \ref{tab:ablation} shows the results against two strong base models: Updn \citep{Anderson_2018_CVPR} and SSL \citep{ijcai2020-151}. SSL is built on UpDn and uses negative sampling to replace images in the triplets of (image, question, answer) to create irrelevant samples.
% \todo{revise this: the difference between what? UpDn and SSL? we never explained SSL  - we need to say a little about it here: SSL is the same as UpDn with the addition of negative sampling? }
We can observe that in both cases, the components in our proposed framework can be beneficial and improve the performance by a reasonable margin.

\input{tables/tab_ablation}

Specifically, the incorporation of generative training for augmentation, i.e. answer generation $q(A | V)$ and question generation $q(Q|V, A)$, improves 1.26 and 2.21 accuracy points on 
% \todo{changed it: either report absolute gains (just number) or relative gains in \%} 
the two base models, respectively.
When introducing the cross-modal joint modelling to provide prior knowledge of multimodal distribution, the performance is further improved by 0.12 and 0.57 accuracy points, respectively.
The overall better results observed with SSL can be attributed to the negative sampling feature in this model.
The most significant improvement across all the categories is observed in ``Num" questions. 
With the generative objective as reward to reweigh the VQA loss on the augmented data, the best performance of our model can achieve is 60.70 accuracy, outperforming the strong SSL model by 3.11 points.

\begin{figure*}[t]
  \centering
  \includegraphics[width=0.99\textwidth]{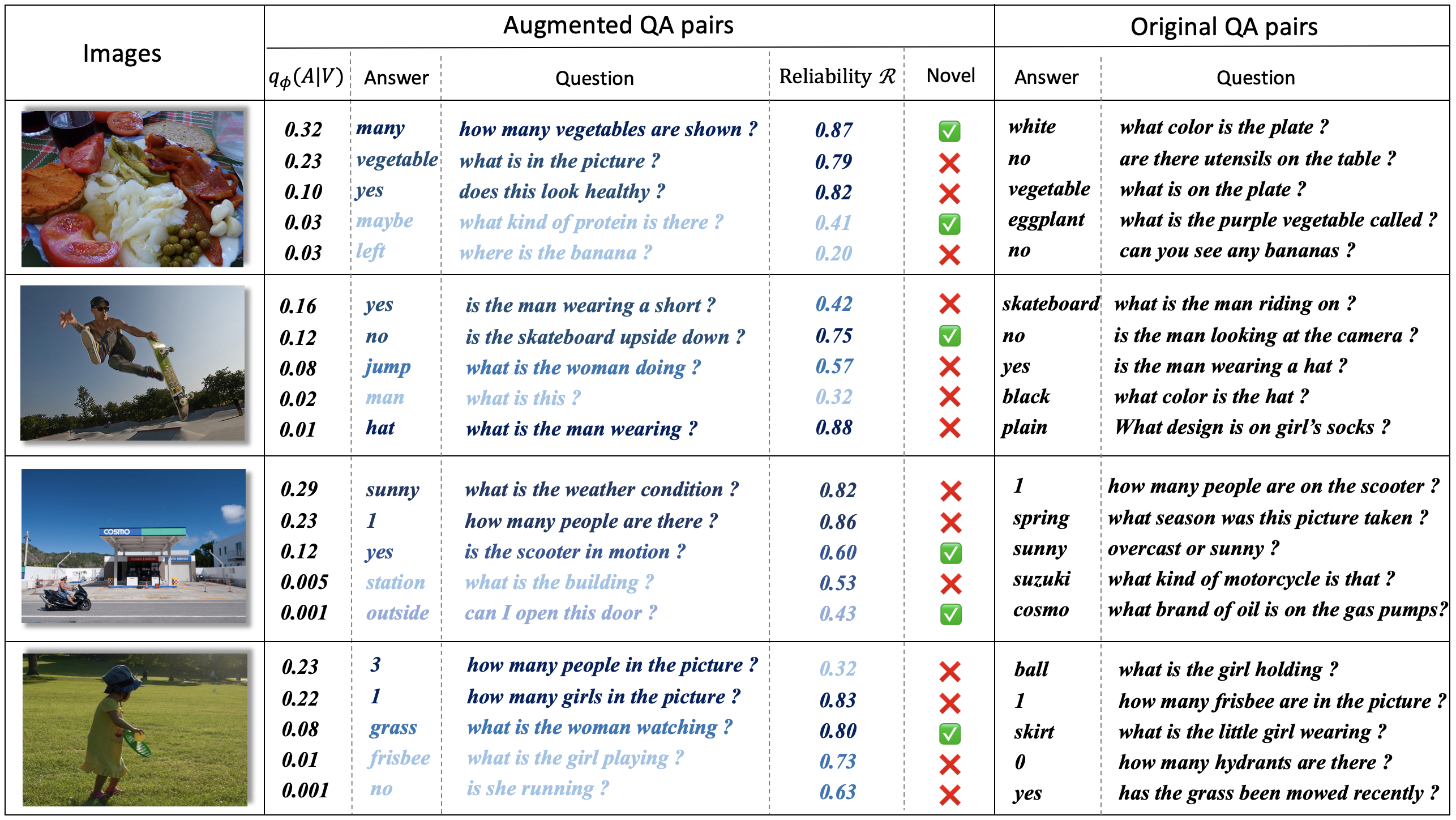}
  \caption{\textbf{Qualitative Analysis:} Examples comparing generated QA pairs against those from the original dataset annotations for the same image. 
  Generated QA pairs are informative if they are out of distribution compared with the original QA pairs.}
  \label{fig:analysis1}
\end{figure*}

\subsection{Examples of Generated QA pairs}

To qualitatively demonstrate the effectiveness of the proposed approach, Figure \ref{fig:analysis1} shows examples of generated QA pairs and compares them with the original ones from the VQA-CP-v2 dataset.

In the four examples, we present five generated QA pairs from the proposed generator and five  QA pairs from the original dataset, for each image.
For the generated QA pairs, we have a column \texttt{$q_{\phi}(A|V)$} showing the probabilities 
% \todo{add equation in column header here}
of sampled answers given only the input image, in descending order.
It can be seen from the \texttt{Answer} column that most of the answers can be directly related to the image, but some are not.
This is because we use ground-truth answers from the original dataset to train the image-answer pipeline, which contains some answers that have high correlations with the questions but not with the image. 
% \todo{are you referring to the answers that are related or the ones that are not? 'this' in 'this is because'?}

In the column of \texttt{Questions}, we show the generated questions that are conditioned on the sampled answers and the image, which have high diversity. 
To quantify diversity, we measure type-token ratio on a per-image basis. 
The average ratio is 0.53, which is -- as expected -- lower than the average ratio for the original QA pairs (0.79), but Figure \ref{fig:analysis1} shows that generated questions are meaningful and related to the answers and the image.

Additionally, the \texttt{Reliability} $\mathcal{R}$ of each QA pair is illustrated, showing that most QA pairs have high reliability and relevance to the image.
Note that high probabilities of answers given an image cannot guarantee that the augmented QA pairs have high correlation with the image.
In the column of \texttt{Novel}, we use green check to indicate the novel generated questions when compared to the set of question in the column of \texttt{Original QA pairs}.
The questions are novel when they convey different information from the original ones.

We also inspect the generated data for a better understanding of the generated QA pairs for different types of questions. 
The improvements over ``NUM'' questions can be due to the answer generator.
Our answer generator can sample diverse possible numbers first given an image and the subsequently generated questions are always ``how many'' questions, which is not always the case for ``Yes/No'' and ``Other'' types.
This provides diverse ``NUM'' QA pairs which are not always ``1'' as the answers in the original dataset.

% \subsubsection{$p_{\theta}(V|Q, A)$}
% \zw{line graph to explain threshold} As we use confidence threshold to filter out generated QA pairs, it is necessary to see the effect of different threshold setups on the contribution of the retrieval pipeline. 
% \subsubsection{Examples of QA pairs?}

\section{Conclusions}
This paper introduces a generative model for cross-modal data augmentation on VQA. 
We learn a generator to generate reliable QA pairs given images under a generative framework.
The augmented QA pairs are trained and evaluated by the generative distribution pretrained on VQA dataset, which are in turn employed in a downstream VQA task with confidence scores to selectively improve the classification performance. 
Without low-level operations or specific heuristics, our proposed model is able to augment unlabelled images with large-scale reasonable QA pairs, which boosts a vanilla model to achieve benchmark results.
The strong generalisation ability of our model open avenues for extension to other multimodal machine learning tasks. 

\section*{Acknowledgements}
This work received support from the MultiMT project (H2020 ERC Starting Grant No. 678017) and the Air Force Office of Scientific Research (under award number FA8655-20-1-7006).

\bibliography{egbib}
\end{document}

%% file: tables/tab_results.tex
\begin{table*}[ht]
	\small
	\begin{center}
		\resizebox{0.95\linewidth}{!}{
			\begin{tabular}{@{}l  cccc c cccc c@{}}
				\toprule
				\multirow{2}{*}{Model}  & \multicolumn{4}{c}{VQA-CP-v2 test (\%) $\uparrow$} & \hphantom & \multicolumn{4}{c}{VQA-v2 test (\%) $\uparrow$} & \\
				\cmidrule{2-5} \cmidrule{7-10}
				& All & Yes/No & Num & Other && All & Yes/No & Num & Other & \\
				\toprule
				% HAN~\cite{malinowski2018learning} & 28.65 & 52.25 & 13.79 & 20.33 && - & - & - & - & -\\
				GVQA~\citep{Agrawal_2018_CVPR} & 31.30 & 57.99 & 13.68 & 22.14 && 48.24 & 72.03 & 31.17 & 34.65 & \\
				RUBi~\citep{NEURIPS2019_51d92be1} & 47.11 & 68.65 & 20.28 & 43.18 && 63.10 & - & - & - & \\ 
				SCR~\citep{NEURIPS2019_33b879e7} & 48.47 & 70.41 & 10.42 & 47.29 && 62.30 & 77.40 & 40.90 & 56.50 & \\
			    LMH~\citep{clark-etal-2019-dont} & 52.45 & 69.81 & 44.46 & 45.54 && 61.64 & 77.85 & 40.03 & 55.04 & \\
			    CSS~\citep{chen2020counterfactual} & 58.95 & 84.37 & 49.42 & 48.21 && 59.91 & 73.25 & 39.77 & 55.11 &  \\
			    \midrule
				\midrule
			    LXMERT~\citep{tan-bansal-2019-lxmert} & 46.23 & 42.84 & 18.91 & 55.51 && \bf 74.16 & \bf 89.31 & \bf 56.85 & \bf 65.14 & \\
				MUTANT~\citep{gokhale-etal-2020-mutant} & \textbf{61.72} & 88.90 & \textbf{49.68} & \textbf{50.78} && 62.56 & 82.07 & 42.52 & 53.28 & \\
				Ours & 46.93 & 43.25 & 20.03 & 54.56 && 59.32 & 80.10 & 40.23 & 49.58 & \\
				
				\midrule
				UpDn~\cite{Anderson_2018_CVPR}  & 41.58 & 43.07 & 13.58 & 48.48 && 63.48 & 81.18 & 42.14 & 55.66 & \\
				UpDn + SSL~\citep{ijcai2020-151} & 57.59 & 86.53 & 29.87 & 50.03 && 63.73 & - & - & - &  \\
				Ours & \underline{60.70} & \bf \underline{89.73} & \underline{45.89} & 48.36 && \underline{64.09} & \underline{81.78} & \underline{44.57} &  \underline{55.84} & \\
				% \midrule
				% LXMERT~\cite{tan2019lxmert} & 46.23 & 42.84 & 18.91 & 55.51 && \textbf{74.16} & \textbf{89.31} & \textbf{56.85} & \textbf{65.14} & 27.97\\
				% LXMERT + MUTANT & \textbf{69.52} & \textbf{93.15} & \textbf{67.17} & \textbf{57.78} &&  \underline{70.24} & \underline{89.01} & \underline{54.21} & \underline{59.96} & \textbf{0.72} \\
				% LXMERT~\cite{tan2019lxmert} (reproduced) &  &  &  &  &&  &  &  &  & \\
				\bottomrule
			\end{tabular}
		} % scale box
	\end{center}
	\caption{Accuracies on VQA-CP-v2 test and VQA-v2 validation sets. ``Ours" represents the final model build on LXMERT/UpDn with augmentation sampler, cross-modal joint distribution and reweighing augmented loss.
	Overall best scores are \textbf{bold}, and our best ones are \underline{underlined}.}
	\label{tab:results}
\end{table*}

%% file: tables/tab_ablation.tex
\begin{table}[t]
	\small
	\begin{center}
		\resizebox{0.5\linewidth}{!}{
            \begin{tabular}{l cccc}
            \toprule
            & \multicolumn{4}{c}{VQA-CP-v2 test (\%)} \\
            \cmidrule{2-5}
            & {\footnotesize All} & {\footnotesize Yes/No} & {\footnotesize Num} &  {\footnotesize Other} \\
            \midrule
            \midrule
            % UpDn & 39.74 & 42.27 & 11.93 & 46.05 \\
            UpDn & 41.58 & 43.07 & 13.58 & 48.48  \\
            + $q_{\phi}(Q, A|V)$ & 42.84 & 44.57 & 24.93 & 47.05 \\
            + $p_{\theta}(V, Q, A)$ & 42.96 & 45.88 & 24.82 & 47.66 \\
            + reweighted & \bf 43.12 & \bf 45.76 & \bf 25.71 & \bf 48.51 \\
            \midrule
            SSL & 57.59 & 86.53 & 29.87 & \bf 50.03 \\
            % + sigmoid &  &  &  &  \\
            + $q_{\phi}(Q, A|V)$ & 59.80 & 89.54 & 43.80 & 48.61 \\
            + $p_{\theta}(V, Q, A)$ & 60.37 & 89.49 & 44.43 & 48.72 \\
            + reweighted & \bf 60.70 & \bf 89.73 & \bf 45.89 & 48.36 \\
            \bottomrule
	        \end{tabular}
    }
    \end{center}
	\caption{Study on the benefits of each component of the proposed approach: augmentation sampler, cross-modal joint distribution, and reweighed loss. The best accuracies are in \textbf{bold}.}
	\label{tab:ablation}
\end{table}

% \begin{table}[h]
% 	\small
% 	\begin{center}
% 		\resizebox{0.7\linewidth}{!}{
%             \begin{tabular}{l c}
%             \toprule
%             & {VQA-CP-v2 test} \\
%             % & {\footnotesize All} \\
%             \midrule
%             \midrule
%             % UpDn & 39.74 & 42.27 & 11.93 & 46.05 \\
%             UpDn & 41.58 \\
%             + $q_{\phi}(Q, A|V)$ & 42.84 \\
%             + $p_{\theta}(V, Q, A)$ & 42.96 \\
%             + reweighted &  \\
%             \midrule
%             SSL & 57.59  \\
%             % + sigmoid &  &  &  &  \\
%             + $q_{\phi}(Q, A|V)$ & 59.80  \\
%             + $p_{\theta}(V, Q, A)$ & 60.37  \\
%             + reweighted & \bf 60.70  \\
%             \bottomrule
% 	        \end{tabular}
%     }
%     \end{center}
% 	\caption{Ablation study to investigate the effect of each component: augmented sampler, cross-modal joint distribution, and reweighted loss. Overall best scores are \textbf{bold}.}
% 	\label{tab:ablation}
% \end{table}